\documentclass[10pt,twocolumn,letterpaper]{article}
\ifdefined\pdfminorversion\pdfminorversion=7\fi
\usepackage[pagenumbers]{cvpr}
\usepackage{amsmath,amssymb,booktabs,tabularx,array,multirow}
\usepackage{colortbl}
\usepackage{graphicx}
\definecolor{oursbg}{gray}{0.925}
\definecolor{methodgreen}{HTML}{287C66}
\definecolor{methodblue}{HTML}{406B9E}
\usepackage{microtype}
\usepackage[most]{tcolorbox}
\usepackage{algorithm,algpseudocode}
\newtcolorbox{promptbox}[1]{enhanced,
  colback=gray!3,colframe=methodblue!70!black,
  colbacktitle=methodblue!12,coltitle=black,
  fonttitle=\bfseries\small,fontupper=\small,title=#1,
  boxrule=0.6pt,arc=1mm,left=2mm,right=2mm,top=1.5mm,bottom=1.5mm}
\definecolor{cvprblue}{rgb}{0.21,0.49,0.74}
\usepackage[breaklinks,colorlinks,allcolors=cvprblue]{hyperref}
\hypersetup{
  pdftitle={SARGO: Agreement-Gated Reinforcement and Suppression via On-Policy Self-Guidance for Visual Hallucination Mitigation},
  pdfauthor={Jinyu Li, Zhiming Zhang, Hao Fang, Hongyao Yu, Bin Chen, Guojun Ma, Shu-Tao Xia}
}
\title{SARGO: Agreement-Gated Reinforcement and Suppression\\via On-Policy Self-Guidance for Visual Hallucination Mitigation}
\author{
\textbf{Jinyu Li}$^{*1,2}$,\quad
\textbf{Zhiming Zhang}$^{*2}$,\quad
\textbf{Hao Fang}$^{*1}$,\quad
\textbf{Hongyao Yu}$^{1}$,\\
\textbf{Bin Chen}$^{\dagger2}$,\quad
\textbf{Albert Ma}$^{1}$,\quad
\textbf{Shu-Tao Xia}$^{1}$\\[3pt]
$^{1}$Tsinghua Shenzhen International Graduate School, Tsinghua University\\
$^{2}$Harbin Institute of Technology, Shenzhen\\[3pt]
{\small\texttt{\{2023311402, 2024312354\}@stu.hit.edu.cn}\quad 
\small\texttt{ffhibnese@gmail.com}}
}
\begin{document}
\maketitle
\begingroup
\renewcommand{\thefootnote}{\fnsymbol{footnote}}
\footnotetext[1]{Equal contribution.}
\footnotetext[2]{Corresponding author.}
\endgroup
\begin{abstract}
Vision-language models remain susceptible to visual hallucinations, often mixing grounded information with unsupported claims in the same response. Response-level reinforcement learning assigns a shared advantage to all tokens in a response, causing negative updates to suppress correct information alongside unsupported claims. To reduce hallucination while preserving useful visual information, we introduce SARGO (Self-guided Advantage-Routed Grounding Optimization), a post-training framework for selective reinforcement and suppression through on-policy self-guidance. Direct and reciprocal weighting gives better-supported tokens stronger reinforcement on positive updates and weaker penalties on negative updates. Agreement between overall and hallucination-specific advantages selects when to apply this guidance. At the response level, a reference-calibrated reward couples factuality with the retention and expansion of grounded information, discouraging lower hallucination scores obtained through omission. Experiments on Qwen3-VL-4B, Qwen3-VL-8B, and Gemma3-4B demonstrate that SARGO outperforms the evaluated mitigation baselines across Obj-Hal, AMBER, THRONE, and SHR evaluations. SARGO achieves these gains while preserving useful visual information and general multimodal capabilities.
\end{abstract}

\section{Introduction}
\label{sec:intro}

Instruction-tuned vision-language models (VLMs) support a broad range of multimodal understanding and reasoning tasks~\cite{liu2023llava,dai2023instructblip,bai2025qwen3vl,gemma2025gemma3}. However, they remain prone to visual hallucinations, producing content unsupported by the input image~\cite{rohrbach2018object,li2023pope}. Hallucinated claims often coexist with accurate observations within the same response. Effective mitigation should suppress unsupported content while preserving grounded visual information.

\begin{figure*}[t]
\centering
\includegraphics[width=\textwidth]{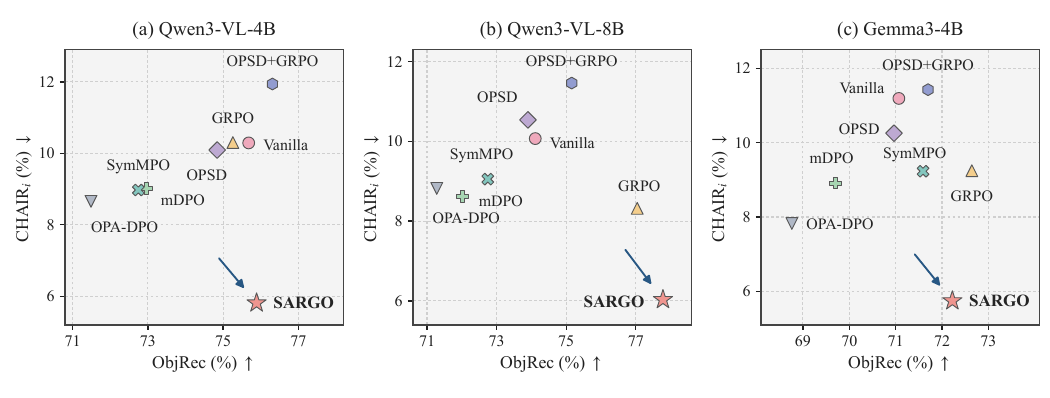}
\caption{\textbf{Object hallucination and information retention on Obj-Hal.} Comparison of SARGO and baselines on (a) Qwen3-VL-4B, (b) Qwen3-VL-8B, and (c) Gemma3-4B. Lower CHAIR$_i$ and higher object recall (ObjRec) indicate better performance. SARGO achieves the lowest CHAIR$_i$ across all three backbones while retaining high object recall.}
\label{fig:teaser}
\end{figure*}

Existing methods~\cite{wang2024mdpo,yang2025opadpo,liu2025symmpo} can suppress hallucinated content at the expense of useful visual information, reducing object coverage. The objective should reward correction together with information retention. Response-level reinforcement learning (RL)~\cite{sun2024llavarlhf}, however, assigns every token the same advantage, suppressing correct visual details alongside hallucinated claims on negative updates.

On-policy self-distillation (OPSD) provides token-level guidance by conditioning the model on privileged context~\cite{zhao2026opsd}. A natural combination adds its teacher--student KL objective to the RL loss. Such distribution matching follows teacher preferences independently of the response advantage, potentially reinforcing tokens that RL seeks to suppress. We instead use teacher support to redistribute reward-driven updates. For hallucination mitigation, this guidance remains valuable even on responses with negative advantages, where grounded details coexist with unsupported claims.

This motivates SARGO (Self-guided Advantage-Routed Grounding Optimization), a post-training framework that uses on-policy self-guidance for selective reinforcement and suppression. The current policy scores its own sampled responses with and without privileged visual context to obtain token-level grounding support. Direct teacher-support weights emphasize well-supported tokens, but applying them to negative advantages would also amplify their penalties. Our central principle is that better-supported tokens should receive stronger reinforcement on positive updates and weaker penalties on negative updates. SARGO therefore uses direct weights for reinforcement and reciprocal weights for suppression.

Teacher guidance should align with the objective driving each update. The overall reward combines factuality with information retention and response quality, potentially yielding an advantage that differs in sign from the hallucination-specific advantage. Our hallucination-aware router applies teacher weighting when the two advantages agree in sign and uses uniform RL weights for all other responses.

Lower hallucination scores can result from omitting useful visual information. We design a reference-calibrated response objective that couples factuality with coverage retention and expansion. Coverage losses are measured against each image's reference coverage, and gains against the remaining headroom. A coverage gate reduces factuality credit when coverage falls, rewarding error reduction together with preserved visual information.

As illustrated in Figure~\ref{fig:teaser}, SARGO achieves the lowest Obj-Hal hallucination rates among evaluated methods on Qwen3-VL-4B, Qwen3-VL-8B, and Gemma3-4B. It consistently outperforms mitigation baselines across hallucination benchmarks and improves coverage over Vanilla while preserving general multimodal capabilities. Ablations establish the complementary contributions of selective reinforcement and suppression, hallucination-aware routing, and reward design. Our contributions are as follows:
\begin{itemize}
    \item We propose SARGO for selective reinforcement and suppression, using direct and reciprocal weighting to modulate token-level update strength and hallucination-aware routing to select when on-policy self-guidance applies.
    \item We introduce a reference-calibrated response objective rewarding factuality alongside retained and newly recovered visual information. Coverage penalties and gains encourage retention and expansion, while a coverage gate ties factuality credit to retained information.
    \item We demonstrate joint hallucination--coverage gains and lower SHR across Qwen3-VL and Gemma3 backbones, with ablations establishing the roles of selective attribution, agreement routing, and reward design.
\end{itemize}

\section{Related Work}
\label{sec:related}

\paragraph{Visual reliability and hallucination mitigation.}
Maintaining consistency between visual evidence and model outputs is central to reliable visual understanding. Adversarial attacks can disrupt image--text alignment~\cite{fang2025one} and manipulate visual predictions~\cite{fang2024clip}. On natural images, VLMs can still generate content unsupported by or inconsistent with the visual evidence, a phenomenon known as visual hallucination~\cite{rohrbach2018object,li2023pope,guan2024hallusionbench}. Existing solutions can be broadly grouped into training-free and training-based approaches. Training-free methods improve grounding through contrastive decoding in VCD~\cite{leng2024vcd}, attention-guided decoding in OPERA~\cite{huang2024opera}, and adaptive focal grounding in HALC~\cite{chen2024halc}. CMI-VLD calibrates decoding through conditional mutual information between visual and textual tokens~\cite{fang2025grounding}. Woodpecker and LURE instead correct unsupported claims in descriptions, improving their consistency with the visual evidence~\cite{yin2024woodpecker,zhou2024lure}.

Training-based approaches strengthen factual supervision through instruction refinement~\cite{liu2024lrv,yu2024hallucidoctor}, human corrections~\cite{yu2024rlhfv}, and model-generated feedback~\cite{yu2025rlaifv}. Building on DPO~\cite{rafailov2023dpo}, HA-DPO and HDPO construct hallucination-aware preferences~\cite{zhao2023beyond,fu2025hdpo}, while OPA-DPO and SymMPO improve preference supervision through reference-policy alignment and symmetric margin regulation, respectively~\cite{yang2025opadpo,liu2025symmpo}. Visual dependence is strengthened through contrastive learning in HACL~\cite{jiang2024hacl}, joint image and response preferences in mDPO~\cite{wang2024mdpo}, and entity-centric alignment in EMPO~\cite{wu2025empo}. Recent studies address hallucination in multimodal reasoning models~\cite{dong2025mirage,fang2026seeing,kong2026reasoning}, multi-image settings~\cite{yang2026looking}, and model merging~\cite{li2026mergeheir}.

\paragraph{Reinforcement learning and on-policy distillation.}
RL post-training optimizes language models using response rewards. RLHF trains a reward model on human preferences and optimizes the policy with PPO~\cite{ouyang2022training,schulman2017ppo}. GRPO estimates advantages from rewards within a response group without a separate critic~\cite{shao2024deepseekmath}. To refine supervision beyond complete responses, Fine-Grained RLHF learns segment-level rewards for factuality, relevance, and completeness~\cite{wu2023finegrained}. Process-supervision methods evaluate intermediate reasoning steps using automated annotations~\cite{luo2024process} or implicit rewards learned from outcome feedback~\cite{cui2025prime}.

Knowledge distillation provides teacher guidance through output or distribution matching. Sequence-level distillation trains on teacher-generated responses~\cite{kim2016sequence}, whereas on-policy distillation queries the teacher on student-generated trajectories. MiniLLM and generalized knowledge distillation use such trajectories for token-level distribution matching~\cite{gu2024minillm,agarwal2024gkd}. OPSD obtains this guidance from the same model under privileged conditioning~\cite{zhao2026opsd}. Distilled RL trains reasoning models on mathematical tasks with binary correctness rewards and incorporates teacher likelihood ratios into policy-gradient weights, retaining uniform weights for nonpositive advantages~\cite{wang2026distilled}. SARGO targets visual hallucination, where unfavorable responses can still contain grounded information. Reciprocal weighting enables selective suppression, while advantage agreement determines when to apply on-policy self-guidance. A reference-calibrated reward complements these token-level updates by jointly evaluating factuality, information retention, and coverage growth.

\section{Method}
\label{sec:method}

\begin{figure*}[t]
\centering
\includegraphics[width=\textwidth]{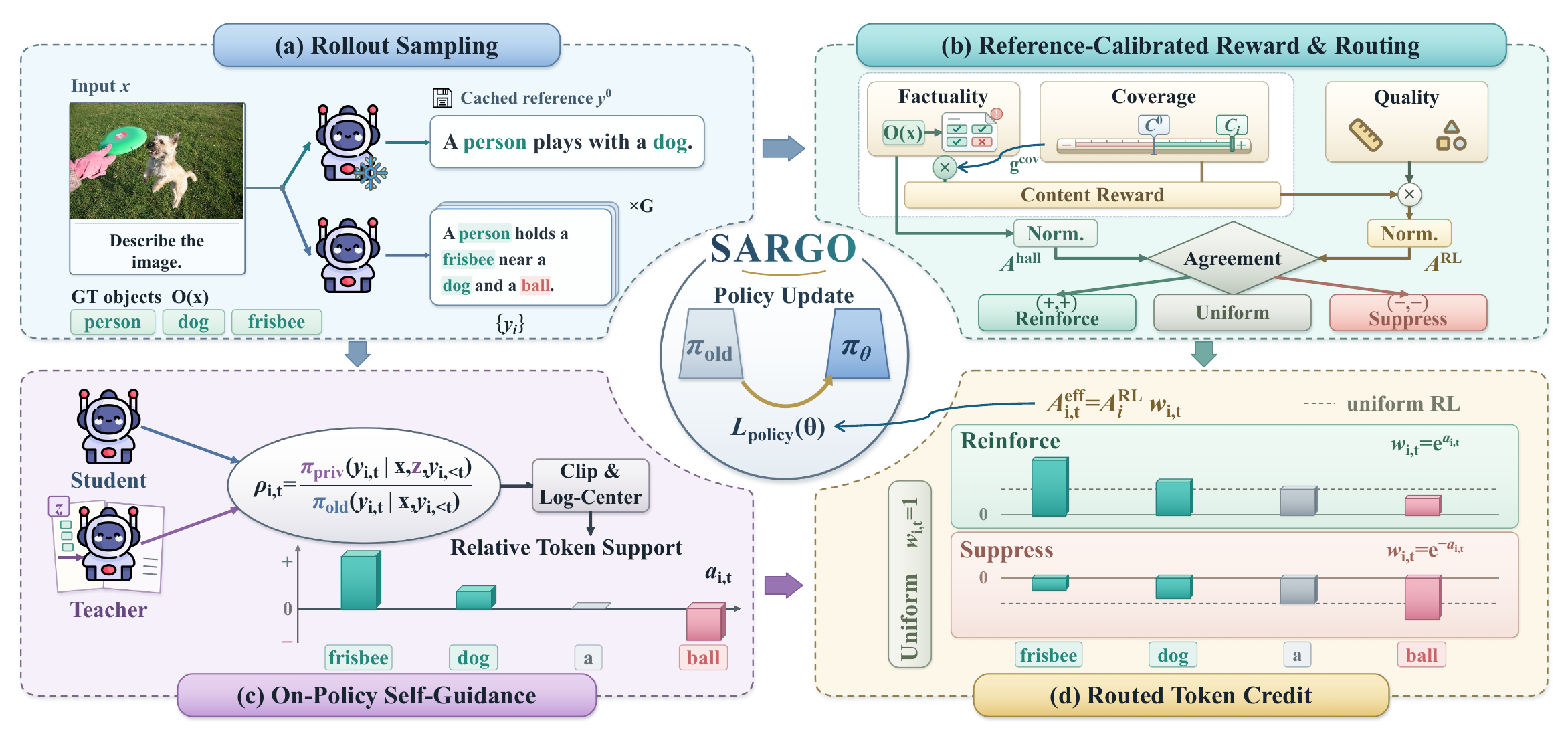}
\caption{\textbf{Overview of SARGO.} The frozen initial policy supplies reference responses for reward calibration, while the current policy scores its own rollouts under ordinary and privileged contexts to obtain token-level grounding support. Agreement between overall and hallucination-specific advantages selects when self-guidance applies. Direct and reciprocal weighting modulate the relative strength of token-level reinforcement and suppression, illustrated by bar magnitude. Neutral routes retain uniform token weights.}
\label{fig:overview}
\end{figure*}

SARGO combines reference-calibrated rewards with selective reinforcement and suppression, as illustrated in Figure~\ref{fig:overview}. Its core principle is to give better-supported tokens stronger reinforcement on positive updates and weaker penalties on negative updates. Given an image and instruction $x$, the rollout policy $\pi_{\mathrm{old}}$ samples $G$ responses $\{y_i\}_{i=1}^{G}$.

\subsection{Reference-Calibrated Response Reward}
\label{sec:reward}
Hallucination reduction should preserve and extend correctly grounded content. We use the frozen initial policy $\pi_0$ to obtain a reference response $y^0$ for each input $x$. Its grounded content provides an image-specific baseline for evaluating retention and coverage gains. Let $\mathcal{O}(x)$ denote annotated object categories and $M(y)$ those extracted from $y$.

\paragraph{Dual-granularity hallucination.}
A small fraction of unsupported categories can still leave false object claims.  We therefore penalize both the fraction of unsupported categories and the presence of any unsupported category:
\begin{equation}
\begin{gathered}
h_i^{\mathrm{obj}}=\frac{|M(y_i)\setminus\mathcal{O}(x)|}{|M(y_i)|},\quad
h_i^{\mathrm{resp}}=\mathbb{I}[|M(y_i)\setminus\mathcal{O}(x)|>0],\\
r_i^{\mathrm{hall}}=1-\alpha h_i^{\mathrm{obj}}-(1-\alpha)h_i^{\mathrm{resp}}.
\end{gathered}
\label{eq:hall}
\end{equation}
The coefficient $\alpha$ balances category-level hallucination severity and response-level hallucination occurrence.

\paragraph{Coverage retention and expansion.}
A response can reduce hallucination by avoiding uncertain object mentions while also omitting correct visual details.  We therefore reward the retention and expansion of object coverage relative to the initial policy's reference response. Let $C_i=|M(y_i)\cap\mathcal{O}(x)|/|\mathcal{O}(x)|$ and $C^0=|M(y^0)\cap\mathcal{O}(x)|/|\mathcal{O}(x)|$ for inputs with nonempty annotations. Relative loss and gain are
\begin{equation}
\Delta_{i,-}=\frac{[C^0-C_i]_+}{C^0},\qquad
\Delta_{i,+}=\frac{[C_i-C^0]_+}{1-C^0}.
\label{eq:covdelta}
\end{equation}
Loss is normalized by reference coverage and gain by the remaining headroom, accounting for differences in initial coverage across images. We define a coverage violation and a retention gate:
\begin{equation}
d_i^{\mathrm{cov}}=\tfrac12+\tfrac12\sqrt{\Delta_{i,-}}-\tfrac12\Delta_{i,+},\quad
g_i^{\mathrm{cov}}=1-\Delta_{i,-}.
\label{eq:cov}
\end{equation}
For equal normalized changes, the square-root penalty gives greater weight to coverage loss than the linear reward gives to coverage expansion, prioritizing the retention of grounded information already captured in the reference response.

\paragraph{Response quality and reward composition.}
To discourage overly brief or repetitive responses during hallucination mitigation, we introduce a response-quality gate. It combines the word counts $L_i$ and $L^0$ of the sampled and reference responses with reference-relative word $n$-gram diversity retention $\delta_i^{\mathrm{div}}$:
\begin{equation}
q_i^{\mathrm{resp}}=
\left(1-\sqrt{\left[1-\frac{L_i}{\beta L^0}\right]_+}\right)
\left(1-\sqrt{1-\delta_i^{\mathrm{div}}}\right).
\label{eq:suff}
\end{equation}
The length factor saturates at $\beta L^0$, while the square roots amplify shortfalls in length and diversity. Appendix~\ref{app:reward_details} defines the reference-relative diversity retention.

The retention gate couples factuality to information preservation: $\widetilde d_i^{\mathrm{hall}}=1-g_i^{\mathrm{cov}}r_i^{\mathrm{hall}}$ discounts factuality when coverage falls below the reference. We combine this violation with coverage and response quality:
\begin{equation}
r_i^{\mathrm{RL}}=q_i^{\mathrm{resp}}\left(1-\sqrt{\frac{(\widetilde d_i^{\mathrm{hall}})^2+(d_i^{\mathrm{cov}})^2}{2}}\right).
\label{eq:reward}
\end{equation}
Squaring emphasizes the larger content violation, while the outer square root preserves the scale of the individual penalties. This encourages the policy to improve factuality while preserving existing grounded information and recovering additional correct visual details.

\subsection{Hallucination-Aware Trajectory Routing}
\label{sec:router}
We obtain teacher guidance from the rollout-policy snapshot conditioned on grounding context $z$, comprising object annotations and reference captions already available for the COCO-Stuff images: $\pi_{\mathrm{priv}}(\cdot\mid x,z,y_{<t})=\pi_{\mathrm{old}}(\cdot\mid x,z,y_{<t})$. Comparing token probabilities with and without $z$ at the same sampled prefix measures how additional grounding evidence changes support for each token in the student's response. The composite reward evaluates coverage and response quality alongside factuality, and its advantage can differ in sign from the hallucination-specific advantage. We apply teacher guidance when these advantages agree in sign, aligning its use with the purpose of the response-level update.  To determine this agreement, we separately normalize $r_i^{\mathrm{RL}}$ and $r_i^{\mathrm{hall}}$ within each rollout group:
\begin{equation}
A_i^k=\frac{r_i^k-\mu_k}{\sigma_k},\qquad
k\in\{\mathrm{RL},\mathrm{hall}\},
\label{eq:adv}
\end{equation}
where $\mu_k$ and $\sigma_k$ are the group mean and standard deviation. The route $s_i$ selects teacher-guided reinforcement, teacher-guided suppression, or uniform attribution:
\begin{equation}
s_i=\begin{cases}
+1,& A_i^{\mathrm{RL}}>0\ \land\ A_i^{\mathrm{hall}}>0,\\
-1,& A_i^{\mathrm{RL}}<0\ \land\ A_i^{\mathrm{hall}}<0,\\
0,&\text{otherwise},
\end{cases}
\label{eq:route}
\end{equation}
A response with above-average factuality can still receive a negative composite advantage because of insufficient coverage or poor response quality. Grounding-specific suppression would then emphasize a different property from the one driving the penalty. Conversely, gains in coverage or response quality can yield a positive composite advantage despite below-average factuality. The composite reward determines whether to reinforce or suppress a response; agreement determines whether grounding-specific teacher support is appropriate for allocating that update across tokens. Neutral routing preserves the response-level update with uniform token weights when the two assessments disagree.

\subsection{Advantage-Aware Token Credit Assignment}
\label{sec:credit}
Teacher support should have complementary effects across update directions: stronger support should amplify reinforcement but attenuate suppression. Direct weighting achieves the former, yet imposes larger penalties on better-supported tokens under a negative advantage. Reciprocal weighting reverses this allocation for suppression.

\paragraph{On-policy teacher support.}
For each token in the student's sampled response, we compare the two distributions at the same prefix to assess its grounding support:
\begin{equation}
\rho_{i,t}=\frac{\pi_{\mathrm{priv}}(y_{i,t}\mid x,z,y_{i,<t})}
{\pi_{\mathrm{old}}(y_{i,t}\mid x,y_{i,<t})}.
\label{eq:teacher-ratio}
\end{equation}
The ratio measures how grounding evidence changes support for a token relative to the policy's ordinary preference for it. This contrast supplies a token-level grounding signal even for responses whose overall advantage is negative. Scoring the student's current rollouts aligns teacher guidance with its evolving generation behavior, enabling selective reinforcement and suppression within its own responses.

Let $T_i$ be the number of tokens in response $y_i$. We stabilize the support ratios by clipping $\rho_{i,t}$ to $[1/c,c]$ with bound $c>1$ and centering its logarithm within each response~\cite{wang2026distilled}:
\begin{equation}
a_{i,t}=\log\bar\rho_{i,t}-\frac1{T_i}\sum_{u=1}^{T_i}\log\bar\rho_{i,u},\quad
\bar\rho_{i,t}=\operatorname{clip}(\rho_{i,t},1/c,c).
\label{eq:norm}
\end{equation}
Log-centering expresses support relative to the response's geometric mean, providing a shared reference for reinforcement and suppression when assigning credit across tokens.

\paragraph{Direct and reciprocal weighting.}
Using the route $s_i$ from Eq.~\ref{eq:route} and centered support $a_{i,t}$, we define the token weight and effective advantage as
\begin{equation}
w_{i,t}=\exp(s_i a_{i,t}),\qquad A_{i,t}^{\mathrm{eff}}=A_i^{\mathrm{RL}}w_{i,t}.
\label{eq:eff}
\end{equation}
On positive routes, direct weighting assigns greater credit to tokens with stronger relative teacher support. On negative routes, reciprocal weighting concentrates penalties on less-supported tokens, allowing unfavorable responses to provide corrective feedback while reducing suppression of their better-supported content. Neutral routes preserve the original RL update with uniform token weights.

\subsection{Policy Optimization}
\label{sec:optimization}
Let $u_{i,t}(\theta)=\pi_\theta(y_{i,t}\mid x,y_{i,<t})/\pi_{\mathrm{old}}(y_{i,t}\mid x,y_{i,<t})$ denote the importance ratio between the updated policy $\pi_\theta$ and the rollout policy. We incorporate the effective token advantages into the clipped group-relative policy objective~\cite{schulman2017ppo,shao2024deepseekmath}:
\begin{equation}
\begin{gathered}
\ell_{i,t}=\min\!\left(u_{i,t}A_{i,t}^{\mathrm{eff}},\operatorname{clip}(u_{i,t},1-\epsilon,1+\epsilon)A_{i,t}^{\mathrm{eff}}\right),\\
\mathcal L_{\mathrm{policy}}(\theta)=-\mathbb{E}\!\left[\frac1G\sum_{i=1}^{G}\frac1{T_i}\sum_{t=1}^{T_i}\ell_{i,t}\right],
\end{gathered}
\label{eq:loss}
\end{equation}
Teacher support enters the objective through reward-conditioned token weights, refining the allocation of an RL update without adding a separate distribution-matching objective. The response advantage determines the update direction, while the support weights determine which tokens receive greater emphasis. This aligns teacher guidance with reward-driven reinforcement and suppression.

\section{Experiments}
\label{sec:experiments}
\begin{table*}[t]
\centering
\caption{Comparison of SARGO and baseline methods on hallucination benchmarks using Qwen3-VL-4B, Qwen3-VL-8B, and Gemma3-4B. THRONE reports classwise $F_{0.5}$ at majority (@5) and unanimous (@9) agreement. All scores are percentages. Bold and underlined values indicate the best and second-best results within each backbone, respectively.}
\label{tab:main_results}
\small
\setlength{\tabcolsep}{3pt}
\begin{tabular}{ccccccccccc}
\toprule
\multirow{2}{*}{Model} & \multicolumn{1}{c|}{\multirow{2}{*}{Method}} & \multicolumn{3}{c|}{Obj-Hal} & \multicolumn{3}{c|}{AMBER} & \multicolumn{2}{c|}{THRONE} & GPT-4-assisted \\
 & \multicolumn{1}{c|}{} & CHAIR$_s\downarrow$ & CHAIR$_i\downarrow$ & \multicolumn{1}{c|}{Recall$\uparrow$} & CHAIR$\downarrow$ & HalRate$\downarrow$ & \multicolumn{1}{c|}{Cover$\uparrow$} & $F_{0.5}$@5$\uparrow$ & \multicolumn{1}{c|}{$F_{0.5}$@9$\uparrow$} & SHR$\downarrow$ \\
\midrule
\multirow{8}{*}{Qwen3-VL-4B} & Vanilla & 55.0 & 10.3 & 75.7 & 8.0 & 60.8 & 74.2 & 74.2 & 81.2 & 34.5 \\
 & OPSD & 55.7 & 10.1 & 74.8 & 7.6 & 60.3 & \underline{74.4} & 73.8 & 81.2 & 33.2 \\
 & GRPO & 45.3 & 10.3 & 75.3 & 7.2 & 46.4 & 71.5 & 76.5 & 86.0 & \underline{25.2} \\
 & OPSD+GRPO & 54.3 & 11.9 & \textbf{76.3} & 9.6 & 60.2 & 73.5 & 74.9 & 83.5 & 30.6 \\
 & mDPO & 48.3 & 9.0 & 73.0 & 6.4 & 44.1 & 68.8 & 76.2 & 85.2 & 30.8 \\
 & OPA-DPO & 44.3 & \underline{8.7} & 71.5 & \underline{5.9} & 38.5 & 66.6 & 75.2 & 84.0 & 31.2 \\
 & SymMPO & \underline{44.0} & 9.0 & 72.7 & 6.0 & \textbf{36.5} & 66.9 & \underline{77.3} & \underline{86.1} & 28.8 \\
\rowcolor{oursbg} \cellcolor{white} & \textbf{SARGO} & \textbf{34.7} & \textbf{5.8} & \underline{75.9} & \textbf{3.9} & \underline{36.7} & \textbf{76.2} & \textbf{82.3} & \textbf{88.0} & \textbf{13.7} \\
\midrule
\multirow{8}{*}{Qwen3-VL-8B} & Vanilla & 54.7 & 10.1 & 74.1 & 7.8 & 60.5 & 74.7 & 73.7 & 81.2 & 30.3 \\
 & OPSD & 56.7 & 10.5 & 73.9 & 7.9 & 62.5 & \underline{75.1} & 72.8 & 81.3 & 29.6 \\
 & GRPO & 44.7 & \underline{8.3} & \underline{77.0} & \underline{6.3} & 43.3 & 74.2 & \underline{77.9} & 84.1 & \underline{25.8} \\
 & OPSD+GRPO & 53.3 & 11.5 & 75.2 & 8.8 & 55.7 & 72.9 & 75.1 & 84.1 & 28.3 \\
 & mDPO & 48.3 & 8.6 & 72.0 & 6.5 & 43.6 & 68.8 & 76.0 & 85.1 & 28.2 \\
 & OPA-DPO & 45.7 & 8.8 & 71.3 & 6.9 & 51.5 & 71.8 & 73.1 & 80.9 & 27.3 \\
 & SymMPO & \underline{43.3} & 9.1 & 72.7 & \underline{6.3} & \textbf{39.3} & 68.2 & 76.9 & \underline{85.7} & 27.8 \\
\rowcolor{oursbg} \cellcolor{white} & \textbf{SARGO} & \textbf{35.3} & \textbf{6.0} & \textbf{77.8} & \textbf{4.7} & \underline{40.1} & \textbf{76.3} & \textbf{80.9} & \textbf{86.4} & \textbf{15.7} \\
\midrule
\multirow{8}{*}{Gemma3-4B} & Vanilla & 54.3 & 11.2 & 71.1 & 6.8 & 54.0 & 68.7 & 71.2 & 85.5 & 24.8 \\
 & OPSD & 51.3 & 10.3 & 71.0 & 6.7 & 52.3 & 68.5 & 71.7 & 85.3 & 23.6 \\
 & GRPO & 47.0 & 9.2 & \textbf{72.6} & 6.6 & 46.7 & \underline{69.5} & 72.7 & 86.9 & \underline{19.7} \\
 & OPSD+GRPO & 52.0 & 11.4 & 71.7 & 9.1 & 59.7 & \underline{69.5} & 71.3 & 84.6 & 22.4 \\
 & mDPO & 47.7 & 8.9 & 69.7 & 6.5 & 52.8 & 69.0 & 72.4 & 85.7 & 22.0 \\
 & OPA-DPO & \underline{43.7} & \underline{7.8} & 68.8 & \underline{5.5} & \underline{43.0} & 65.7 & \underline{73.4} & \underline{87.0} & 21.6 \\
 & SymMPO & 46.0 & 9.2 & 71.6 & 6.6 & 53.2 & 68.7 & 72.5 & 86.1 & 22.6 \\
\rowcolor{oursbg} \cellcolor{white} & \textbf{SARGO} & \textbf{30.7} & \textbf{5.7} & \underline{72.2} & \textbf{5.2} & \textbf{35.6} & \textbf{70.2} & \textbf{77.0} & \textbf{88.3} & \textbf{17.8} \\
\bottomrule
\end{tabular}
\end{table*}

\subsection{Experimental Setup}
\noindent\textbf{Models and datasets.}
We conduct experiments on Qwen3-VL-4B-Instruct, Qwen3-VL-8B-Instruct~\cite{bai2025qwen3vl}, and Gemma3-4B-IT~\cite{gemma2025gemma3}. All three models are trained on the same 10,000 COCO-Stuff images with accompanying visual grounding information~\cite{lin2014coco,caesar2018cocostuff}. The training set is strictly disjoint from the image sets used for evaluation across all benchmarks.

\noindent\textbf{Evaluation benchmarks.}
We assess object hallucination on Obj-Hal using response-level CHAIR$_s$, object-level CHAIR$_i$~\cite{rohrbach2018object}, and object recall. On AMBER~\cite{wang2023amber}, we report CHAIR, hallucination rate (HalRate), and coverage (Cover). THRONE~\cite{kaul2024throne} evaluates object-presence decisions with classwise precision, recall, and $F_{0.5}$; @5 and @9 denote majority and unanimous agreement among nine votes. We further use GPT-4-assisted evaluation~\citep{zhao2023beyond} to assess fine-grained hallucinations in generated image descriptions through the Sentence-level Hallucination Ratio (SHR). General multimodal capabilities on Qwen3-VL are evaluated using the total MME score~\cite{fu2025mme} and MMStar accuracy~\cite{chen2024mmstar}.

\noindent\textbf{Baselines.}
We compare with the original backbone (Vanilla), OPSD~\cite{zhao2026opsd}, GRPO~\cite{shao2024deepseekmath}, and OPSD+GRPO. OPSD applies on-policy teacher--student distribution matching. GRPO uses the same response reward in Eq.~\ref{eq:reward} with uniform token weights. OPSD+GRPO combines the two losses additively. We also compare with strong training-based baselines, including mDPO~\cite{wang2024mdpo}, OPA-DPO~\cite{yang2025opadpo}, and SymMPO~\cite{liu2025symmpo}, trained following their original settings.

\noindent\textbf{Implementation details.}
We train all three backbones for one epoch using AdamW with a learning rate of $10^{-6}$ and an effective batch of 512 response sequences. Each rollout samples $G=8$ responses per image with temperature $1.0$ and top-$p=0.95$. Reference descriptions for reward calibration are generated by the initial backbone and cached before training. We set the hallucination mixture coefficient $\alpha=0.7$, reference scaling factor $\beta=0.35$, teacher-ratio clipping bound $c=3$, and policy clipping margin $\epsilon=0.2$. Diversity retention uses word 1--4-grams. To reduce sampling-induced variance and maintain consistent comparisons across benchmarks, we use greedy decoding for all model responses during evaluation. Appendix~\ref{app:settings} provides additional training details and benchmark descriptions.

\subsection{Hallucination Mitigation and Information Retention}
\noindent\textbf{Factuality and coverage improve together.}
As shown in Table~\ref{tab:main_results}, SARGO achieves lower AMBER CHAIR and higher coverage than all evaluated baselines across all three backbones. On Qwen3-VL-4B, for example, CHAIR decreases from OPA-DPO's $5.9$ to $3.9$, while coverage increases from $66.6$ to $76.2$. On Qwen3-VL-4B and Qwen3-VL-8B, SymMPO achieves a similar HalRate to SARGO but has substantially higher CHAIR and lower coverage. Strong training-based baselines such as OPA-DPO and SymMPO reduce hallucination on Qwen3-VL with coverage below Vanilla, suggesting gains partly at the expense of grounded information. SARGO lowers CHAIR while raising coverage above Vanilla across all three backbones. The Obj-Hal results show a similar pattern: SARGO obtains the lowest CHAIR$_s$ and CHAIR$_i$ across all three backbones, with object recall ranking first on Qwen3-VL-8B and second on Qwen3-VL-4B and Gemma3-4B. Relative to Vanilla, CHAIR$_i$ decreases by $40.6\%$--$49.1\%$ across the three backbones. On Gemma3-4B, SARGO lowers OPA-DPO's CHAIR$_i$ from $7.8$ to $5.7$ and increases object recall from $68.8$ to $72.2$.

\begin{figure*}[t]
\centering
\includegraphics[width=\textwidth]{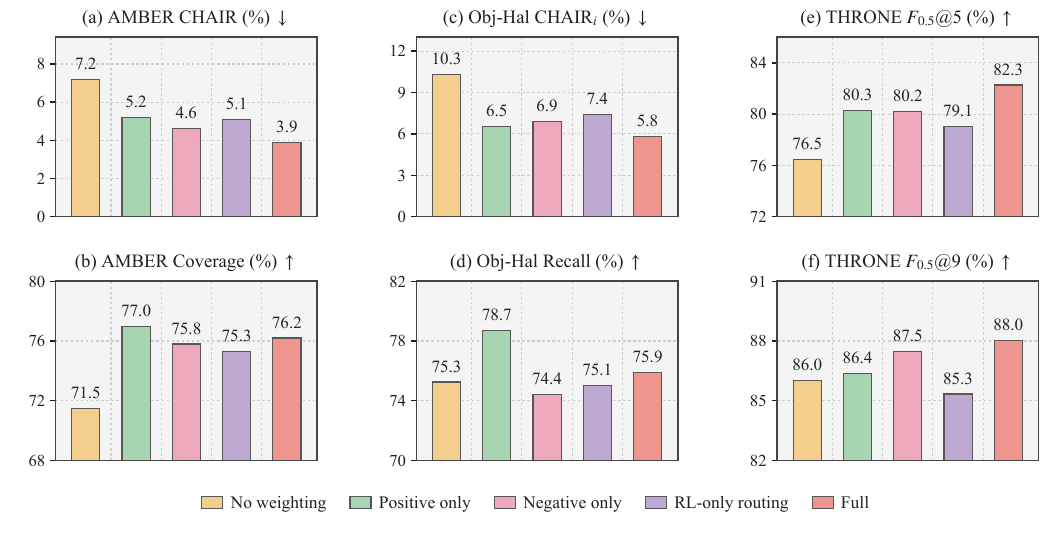}
\caption{\textbf{Effects of token weighting and advantage routing on Qwen3-VL-4B.} All variants share the same response reward. No weighting uses uniform token weights. Positive only and Negative only apply teacher weighting in one branch, retaining uniform weights in the other. RL-only routing uses the overall advantage alone to select the weighting branch. Full denotes SARGO.}
\label{fig:ablation}
\end{figure*}
\begin{table*}[t]
\centering
\caption{Ablation study of SARGO's response reward on Qwen3-VL-4B. Each variant removes one component from the full reward.}
\label{tab:reward_ablation_results}
\small
\setlength{\tabcolsep}{3.2pt}
\begin{tabular}{ccccccccc}
\toprule
\multicolumn{1}{c|}{\multirow{2}{*}{Method}} & \multicolumn{3}{c|}{Obj-Hal} & \multicolumn{3}{c|}{AMBER} & \multicolumn{2}{c}{THRONE} \\
\multicolumn{1}{c|}{} & CHAIR$_s\downarrow$ & CHAIR$_i\downarrow$ & \multicolumn{1}{c|}{Recall$\uparrow$} & CHAIR$\downarrow$ & HalRate$\downarrow$ & \multicolumn{1}{c|}{Cover$\uparrow$} & $F_{0.5}$@5$\uparrow$ & $F_{0.5}$@9$\uparrow$ \\
\midrule
\rowcolor{oursbg} \textbf{SARGO} & 34.7 & 5.8 & 75.9 & 3.9 & 36.7 & 76.2 & 82.3 & 88.0 \\
w/o response quality & 11.0 & 3.0 & 72.7 & 1.7 & 6.8 & 58.0 & 82.9 & 87.8 \\
w/o hallucination & 51.0 & 10.4 & 79.9 & 5.6 & 49.5 & 77.2 & 79.4 & 85.8 \\
w/o coverage & 30.7 & 5.4 & 68.3 & 4.1 & 35.3 & 62.3 & 81.0 & 86.4 \\
w/o coverage gate & 35.0 & 6.0 & 70.4 & 4.1 & 37.5 & 69.8 & 80.8 & 86.5 \\
\bottomrule
\end{tabular}
\end{table*}

\noindent\textbf{Integrating on-policy self-guidance.}
Under the same response reward, adding OPSD's distribution-matching loss to GRPO increases object hallucination on all three backbones. AMBER CHAIR rises from $7.2$ to $9.6$ on Qwen3-VL-4B, from $6.3$ to $8.8$ on Qwen3-VL-8B, and from $6.6$ to $9.1$ on Gemma3-4B. Obj-Hal CHAIR$_i$ worsens in all three cases as well. Both hallucination metrics are also higher than Vanilla, showing that additive distribution matching fails to preserve GRPO's factuality gains in this setting. SARGO improves the joint outcome, achieving lower CHAIR and higher AMBER coverage than OPSD+GRPO across all three backbones. On Qwen3-VL-8B, it reduces OPSD+GRPO's CHAIR$_i$ from $11.5$ to $6.0$ while increasing recall from $75.2$ to $77.8$. On both 4B backbones, it approximately halves OPSD+GRPO's CHAIR$_i$ while maintaining comparable object recall. AMBER coverage increases over OPSD+GRPO by $2.7$, $3.4$, and $0.7$ points on Qwen3-VL-4B, Qwen3-VL-8B, and Gemma3-4B, respectively. These comparisons highlight the importance of how teacher guidance enters RL: SARGO uses the response advantage to determine the update direction and teacher support to distribute reinforcement and suppression across tokens.

\noindent\textbf{Grounding gains extend to THRONE.}
SARGO achieves the highest reported THRONE $F_{0.5}$ at both voting thresholds across all three backbones. At majority agreement, it exceeds SymMPO, the strongest Qwen3-VL-4B baseline, by $5.0$ points. On Gemma3-4B, the gains over OPA-DPO are $3.6$ points at @5 and $1.3$ points at @9, showing improved object-presence judgments under both agreement criteria.

\noindent\textbf{Improved sentence-level factuality.}
As shown in Table~\ref{tab:main_results}, SARGO achieves the lowest SHR among all evaluated methods across the three backbones. Compared with Vanilla, SHR decreases from $34.5$ to $13.7$ on Qwen3-VL-4B, from $30.3$ to $15.7$ on Qwen3-VL-8B, and from $24.8$ to $17.8$ on Gemma3-4B. These results extend the observed gains from object-level grounding to sentence-level factuality.

\begin{figure*}[t]
\centering
\includegraphics[width=\textwidth]{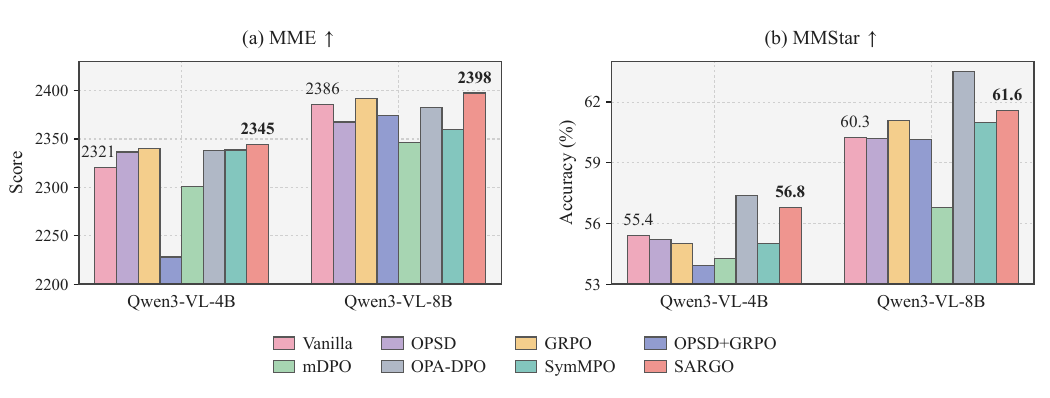}
\caption{\textbf{General multimodal performance on Qwen3-VL-4B and Qwen3-VL-8B.} Comparison on (a) MME total score and (b) MMStar accuracy (\%). SARGO improves both metrics over Vanilla at both scales.}
\label{fig:general_capability}
\end{figure*}

\subsection{Routing and Selective Token Attribution}
\label{sec:ablation}
Figure~\ref{fig:ablation} compares five Qwen3-VL-4B settings under the same reference-calibrated reward. No weighting isolates token attribution; the single-direction variants retain ordinary RL in the opposite direction. RL-only routing retains both branches and tests whether the composite advantage alone can select the teacher's role.

\noindent\textbf{Token attribution improves response-level optimization.}
With the same response reward, Full lowers GRPO's Obj-Hal CHAIR$_i$ from $10.3$ to $5.8$ and AMBER CHAIR from $7.2$ to $3.9$, while coverage rises from $71.5$ to $76.2$. Token attribution thus improves both factuality and information retention beyond response-level optimization.

\noindent\textbf{Negative attribution strengthens hallucination suppression.}
Adding the negative branch to positive-only weighting lowers AMBER CHAIR from $5.2$ to $3.9$ and HalRate from $47.9$ to $36.7$, with coverage changing from $77.0$ to $76.2$. Since the positive branch and response reward are retained, this comparison isolates the benefit of allocating negative credit according to teacher support.

\noindent\textbf{Positive attribution preserves grounded information.}
Relative to negative-only weighting, adding the positive branch lowers Obj-Hal CHAIR$_i$ from $6.9$ to $5.8$ while increasing recall from $74.4$ to $75.9$. AMBER CHAIR and coverage also improve. The single-branch variants expose a tradeoff: positive-only weighting retains more objects, whereas negative-only weighting yields lower AMBER CHAIR. Their combination achieves lower CHAIR scores than either variant, with positive attribution recovering grounded information beyond negative-only updates.

THRONE provides a complementary view of the two branches. Positive-only weighting has a slightly higher $F_{0.5}$ than negative-only weighting at majority agreement, whereas negative-only weighting performs better at unanimous agreement. Their combination achieves the highest score at both thresholds, reaching $82.3$ at @5 and $88.0$ at @9. The combined benefit extends beyond the CHAIR--recall comparison to object-presence judgments, with neither single branch matching the full model across the two agreement criteria.

\noindent\textbf{Advantage agreement selects when to apply guidance.}
RL-advantage-only routing retains both direct and reciprocal weights, replacing Eq.~\ref{eq:route} with $s_i^{\mathrm{RL}}=\operatorname{sign}(A_i^{\mathrm{RL}})$. Restoring hallucination-advantage agreement improves every reported Obj-Hal and AMBER metric and both THRONE scores. CHAIR$_i$ decreases from $7.4$ to $5.8$ and AMBER CHAIR from $5.1$ to $3.9$, relative reductions of $21.6\%$ and $23.5\%$. THRONE $F_{0.5}$ rises from $79.1$ to $82.3$ at @5 and from $85.3$ to $88.0$ at @9. With identical weighting formulas, these gains isolate the value of requiring factuality agreement when allocating teacher-guided updates.

\subsection{Reward Design and Information Retention}
\label{sec:hacking}
Table~\ref{tab:reward_ablation_results} isolates reward components on Qwen3-VL-4B. Coverage rewards retention and gains relative to each image's reference; removing it lowers AMBER coverage from $76.2$ to $62.3$ and Obj-Hal recall from $75.9$ to $68.3$. The coverage gate makes factuality rewards depend on retained information: without it, AMBER coverage falls to $69.8$. Both removals also raise AMBER CHAIR from $3.9$ to $4.1$. Their identical CHAIR scores accompany different losses of correct information, illustrating why hallucination and coverage need to be assessed together.

The two coverage mechanisms are complementary. Once a response matches the reference coverage, the retention gate saturates, while the explicit coverage term continues to reward further correct discoveries through its gain component. Removing this term preserves the gate's protection against information loss but eliminates the incentive for coverage expansion, consistent with the larger drops in AMBER coverage and Obj-Hal recall.

The hallucination term suppresses unsupported content. Removing it raises AMBER coverage to $77.2$ and Obj-Hal recall to $79.9$, but AMBER CHAIR rises to $5.6$ and Obj-Hal CHAIR$_i$ to $10.4$. SARGO combines AMBER CHAIR $3.9$ with coverage $76.2$ and achieves the highest THRONE $F_{0.5}$ of $88.0$ at @9, rewarding informative, grounded responses.

Response quality, $q_i^{\mathrm{resp}}$, favors informative responses. Removing this gate produces extremely short answers with little useful visual content: AMBER CHAIR drops to $1.7$ while coverage falls to $58.0$. The lower hallucination score accompanies substantial content omission, making the response-quality gate important for retaining informative outputs.

\subsection{Preserving General Multimodal Capabilities}
Figure~\ref{fig:general_capability} shows that SARGO achieves the highest MME scores at both Qwen3-VL scales, improving Vanilla by $24$ points on 4B and $12$ on 8B. MMStar accuracy rises from $55.4$ to $56.8$ on 4B and from $60.3$ to $61.6$ on 8B, ranking second at both scales behind OPA-DPO. On 8B, SARGO achieves MME $2398$ and MMStar $61.6$, exceeding GRPO's $2392$ and $61.1$. GRPO falls below Vanilla on 4B MMStar, while OPA-DPO leads on MMStar but falls below Vanilla on 8B MME. SARGO is the only evaluated method to improve over Vanilla on both benchmarks at both Qwen3-VL scales. These results show that stronger hallucination suppression can coexist with preserved general multimodal capabilities.

\par
\section{Conclusion}
This work investigates visual hallucination mitigation through response-level reward design and token-level credit assignment. We propose SARGO, which uses on-policy teacher support to selectively reinforce grounded content and suppress unsupported claims. Direct and reciprocal weighting adapt this guidance to positive and negative updates, while hallucination-aware routing aligns its use with the purpose of each update. A reference-calibrated reward complements these updates by jointly valuing factuality, information retention, and coverage gains. Experiments on Qwen3-VL-4B, Qwen3-VL-8B, and Gemma3-4B demonstrate joint improvements in hallucination mitigation and information retention. Evaluations on Qwen3-VL further show improved general multimodal capabilities. These findings highlight the importance of coordinating selective reinforcement and suppression with reward objectives that preserve and extend grounded visual information.

{\small
\bibliographystyle{ieeenat_fullname}
\bibliography{references}
}
\clearpage
\appendix
\raggedbottom
\section{Experimental Details}
\label{app:settings}

\subsection{Training Configuration}
We train Qwen3-VL-4B, Qwen3-VL-8B~\cite{bai2025qwen3vl}, and Gemma3-4B~\cite{gemma2025gemma3} on the same 10,000 COCO image--instruction examples with accompanying grounding annotations~\cite{lin2014coco}. COCO-Stuff~\cite{caesar2018cocostuff} supplies semantic labels for scene and background regions, which complement object annotations and reference captions in the teacher's grounding context. Training and evaluation use disjoint image sets. All three backbones are trained for one epoch using AdamW with a learning rate of $10^{-6}$ and an effective batch of 512 response sequences. Each rollout samples $G=8$ responses per image with temperature $1.0$ and top-$p=0.95$. Reference responses from the initial model are cached before training.

\subsection{Evaluation Benchmarks}
\paragraph{Object HalBench.}
Object HalBench (Obj-Hal) assesses whether objects mentioned in generated descriptions are supported by image annotations. We report CHAIR$_s$ and CHAIR$_i$~\cite{rohrbach2018object}, which measure the fraction of responses containing hallucinated objects and the fraction of object mentions that are hallucinated, respectively. Object recall also measures the recovery of annotated objects.

\paragraph{AMBER.}
AMBER~\cite{wang2023amber} evaluates hallucinations in generative and discriminative settings. We use its generative setting and report CHAIR, HalRate, and Cover. CHAIR measures the proportion of hallucinated objects, while HalRate measures how often a response contains hallucination. Cover measures the recovery of annotated objects, capturing the amount of grounded information retained in a response.

\paragraph{THRONE.}
Following the original THRONE protocol~\cite{kaul2024throne}, object-presence predictions are assessed against COCO 2017 validation annotations covering 80 object categories. Object-presence judgments are obtained from FLAN-T5 base, large, and XL with three probes each, yielding nine votes. We report classwise $F_{0.5}$ at majority (@5) and unanimous (@9) agreement. This score combines precision and recall with greater emphasis on precision than recall. Scores are macro-averaged over object categories.

\paragraph{GPT-4-assisted evaluation.}
Following~\citet{zhao2023beyond}, we assess fine-grained hallucinations using Visual Genome annotations~\cite{krishna2017visual} and GPT-4o-mini. The evaluator compares generated descriptions with the annotated visual content and identifies hallucinated statements. We report the Sentence-level Hallucination Ratio (SHR) as a percentage. This evaluation complements object-based metrics by assessing factual consistency at the sentence level.

\paragraph{General multimodal benchmarks.}
MME~\cite{fu2025mme} covers ten perception tasks and four cognition tasks, ranging from object counting, spatial localization, and text recognition to commonsense reasoning and numerical calculation. Paired yes/no questions for each image provide a consistent answer format across these tasks. We report the total score combining perception and cognition. MMStar~\cite{chen2024mmstar} contains 1,500 multiple-choice questions selected to require visual evidence. It spans six core capabilities: coarse perception, fine-grained perception, instance reasoning, logical reasoning, science and technology, and mathematics, with 18 finer-grained evaluation dimensions. We report overall accuracy across these questions. Together, these benchmarks assess whether hallucination-focused training preserves visual perception and reasoning for broader multimodal understanding.

\begin{algorithm*}[t]
\caption{Core policy optimization in SARGO}
\label{alg:sargo_training}
\small
\begin{algorithmic}[1]
\Require Training set $\mathcal D=\{(x,\mathcal O(x),z)\}$; initial policy $\pi_0$; group size $G$; parameters $\alpha,\beta,c,\epsilon$
\Ensure Trained policy $\pi_\theta$
\State Freeze $\pi_0$ and initialize $\pi_\theta\gets\pi_0$
\State Cache reference $y^0(x)$ and its calibration statistics from $\pi_0$ for every training input $x$
\For{each training batch $\mathcal B\subset\mathcal D$}
    \State $\pi_{\mathrm{old}}\gets\pi_\theta$
    \For{each $(x,\mathcal O(x),z)\in\mathcal B$}
        \State Sample $G$ student responses $\{y_i\}_{i=1}^{G}\sim\pi_{\mathrm{old}}(\cdot\mid x)$
        \State Compute hallucination rewards $r_i^{\mathrm{hall}}$ from $y_i$ and $\mathcal O(x)$ for all $i$
        \State Calibrate coverage and response quality against cached $y^0(x)$; compose rewards $r_i^{\mathrm{RL}}$
        \State $A_i^k\gets(r_i^k-\mu_k)/\sigma_k$ for $k\in\{\mathrm{RL},\mathrm{hall}\}$, using this input's response group
        \For{each response $y_i$}
            \State $s_i\gets\operatorname{sign}(A_i^{\mathrm{RL}})\,\mathbb{I}[A_i^{\mathrm{RL}}A_i^{\mathrm{hall}}>0]$
            \State $p_{i,t}^{\mathrm{ord}}\gets\pi_{\mathrm{old}}(y_{i,t}\mid x,y_{i,<t})$,\quad
                   $p_{i,t}^{\mathrm{priv}}\gets\pi_{\mathrm{old}}(y_{i,t}\mid x,z,y_{i,<t})$ for all tokens $t$
            \State $\bar\rho_{i,t}\gets\operatorname{clip}(p_{i,t}^{\mathrm{priv}}/p_{i,t}^{\mathrm{ord}},1/c,c)$
            \State $a_{i,t}\gets\log\bar\rho_{i,t}-T_i^{-1}\sum_{u=1}^{T_i}\log\bar\rho_{i,u}$
            \State $w_{i,t}\gets\exp(s_i a_{i,t})$,\quad $A_{i,t}^{\mathrm{eff}}\gets A_i^{\mathrm{RL}}w_{i,t}$
        \EndFor
    \EndFor
    \State Compute the clipped policy loss $\mathcal L_{\mathrm{policy}}$ with effective token advantages and ratio $\pi_\theta/\pi_{\mathrm{old}}$
    \State Update $\theta$ with AdamW on $\mathcal L_{\mathrm{policy}}$
\EndFor
\State \Return $\pi_\theta$
\end{algorithmic}
\end{algorithm*}

\begin{figure}[!t]
\begin{promptbox}{Privileged Grounding Prompt}
[Privileged Grounding Evidence]\\
Use the following information only as private grounding evidence for answering the original request.

\smallskip\textbf{Verified COCO Object Annotations}\\
\texttt{\{object labels\}}

\smallskip\textbf{Verified Scene and Background Regions}\\
\texttt{\{selected scene labels, when available\}}

\smallskip\textbf{Human Observations of the Same Image}\\
1. \texttt{\{reference caption\}}\\
\ldots

Answer the original request directly.
\end{promptbox}
\caption{\textbf{Teacher prompt for on-policy self-guidance.} Image-specific object labels, scene cues, and reference captions provide grounding context for scoring the current policy's sampled responses.}
\label{fig:teacher_prompt}
\end{figure}

\section{Training Procedure and Prompts}
\label{app:teacher}
\subsection{On-Policy Training}
Algorithm~\ref{alg:sargo_training} summarizes the core policy optimization. The frozen initial policy $\pi_0$ supplies cached references for reward calibration. The current policy snapshot $\pi_{\mathrm{old}}$ is refreshed for each training batch and serves both as the rollout policy and as the teacher under privileged conditioning.

\subsection{Grounding Context and Teacher Prompt}
The teacher receives the student's image and instruction together with COCO object labels, human reference captions, and available COCO-Stuff scene cues. Figure~\ref{fig:teacher_prompt} shows the grounding prompt, whose placeholders are filled with the evidence for each image. The scene-related portion is omitted when that evidence is unavailable.

\paragraph{Object extraction.}
Object mentions are mapped to distinct COCO categories using a category lexicon and lemmatization. The reward annotation set combines instance labels with recognized categories from human captions.

\section{Response-Quality Gate}
\label{app:reward_details}

The response-quality gate combines sufficient content with diversity retention relative to the initial policy. Let $\mathcal G_n(y)$ be the multiset of word $n$-grams in response $y$. The fraction of distinct $n$-grams is
\begin{equation}
V_i^{(n)}=\frac{|\operatorname{Unique}(\mathcal G_n(y_i))|}{|\mathcal G_n(y_i)|}.
\end{equation}
For each $n\in\{1,2,3,4\}$, diversity retention relative to the cached reference $y^0$ is
\begin{equation}
\delta_i^{(n)}=\begin{cases}
1,&|\mathcal G_n(y^0)|=0,\\
\min\!\left(\dfrac{V_i^{(n)}}{V_0^{(n)}},1\right),&\text{otherwise}.
\end{cases}
\end{equation}
Orders absent from the reference contribute neutral retention. Combining the four orders gives $\delta_i^{\mathrm{div}}=\prod_{n=1}^{4}\delta_i^{(n)}$, which enters the response-quality gate as
\begin{equation}
\begin{gathered}
q_i^{\mathrm{resp}}=g_i^{\mathrm{len}}\left(1-\sqrt{1-\delta_i^{\mathrm{div}}}\right),\\
g_i^{\mathrm{len}}=1-\sqrt{\left[1-\frac{L_i}{\beta L^0}\right]_+}.
\end{gathered}
\end{equation}
Combining multiple orders accounts for repeated words and longer recurring phrases. This gate reduces the reward of repetitive or insufficient responses, complementing the factuality and coverage terms that evaluate their visual content.

\section{General Multimodal Capabilities}
\label{app:additional_results}

Table~\ref{tab:general_results} gives the numerical results underlying the capability comparison in Figure~\ref{fig:general_capability}. SARGO ranks first on MME and second on MMStar at both model scales, improving both metrics over Vanilla.
\begin{table}[htbp]
\centering
\caption{General multimodal results on MME and MMStar. MME reports total scores and MMStar reports accuracy (\%). Bold and underlined values denote the best and second-best results within each backbone, respectively.}
\label{tab:general_results}
\small
\setlength{\tabcolsep}{3pt}
\begin{tabular}{ccccc}
\toprule
& \multicolumn{2}{c}{Qwen3-VL-4B} & \multicolumn{2}{c}{Qwen3-VL-8B} \\
\cmidrule(lr){2-3}\cmidrule(lr){4-5}
Method & MME$\uparrow$ & MMStar$\uparrow$ & MME$\uparrow$ & MMStar$\uparrow$ \\
\midrule
Vanilla & 2321 & 55.4 & 2386 & 60.3 \\
OPSD & 2337 & 55.2 & 2368 & 60.2 \\
GRPO & \underline{2340} & 55.0 & \underline{2392} & 61.1 \\
OPSD+GRPO & 2228 & 53.9 & 2374 & 60.1 \\
mDPO & 2301 & 54.3 & 2346 & 56.8 \\
OPA-DPO & 2338 & \textbf{57.4} & 2382 & \textbf{63.5} \\
SymMPO & 2339 & 55.0 & 2359 & 61.0 \\
\rowcolor{oursbg} \textbf{SARGO} & \textbf{2345} & \underline{56.8} & \textbf{2398} & \underline{61.6} \\
\bottomrule
\end{tabular}
\end{table}

\section{Case Study}
\label{sec:case-study}

\begin{figure*}[p]
    \centering
    \includegraphics[
        page=1,
        width=\textwidth,
        height=0.43\textheight,
        keepaspectratio
    ]{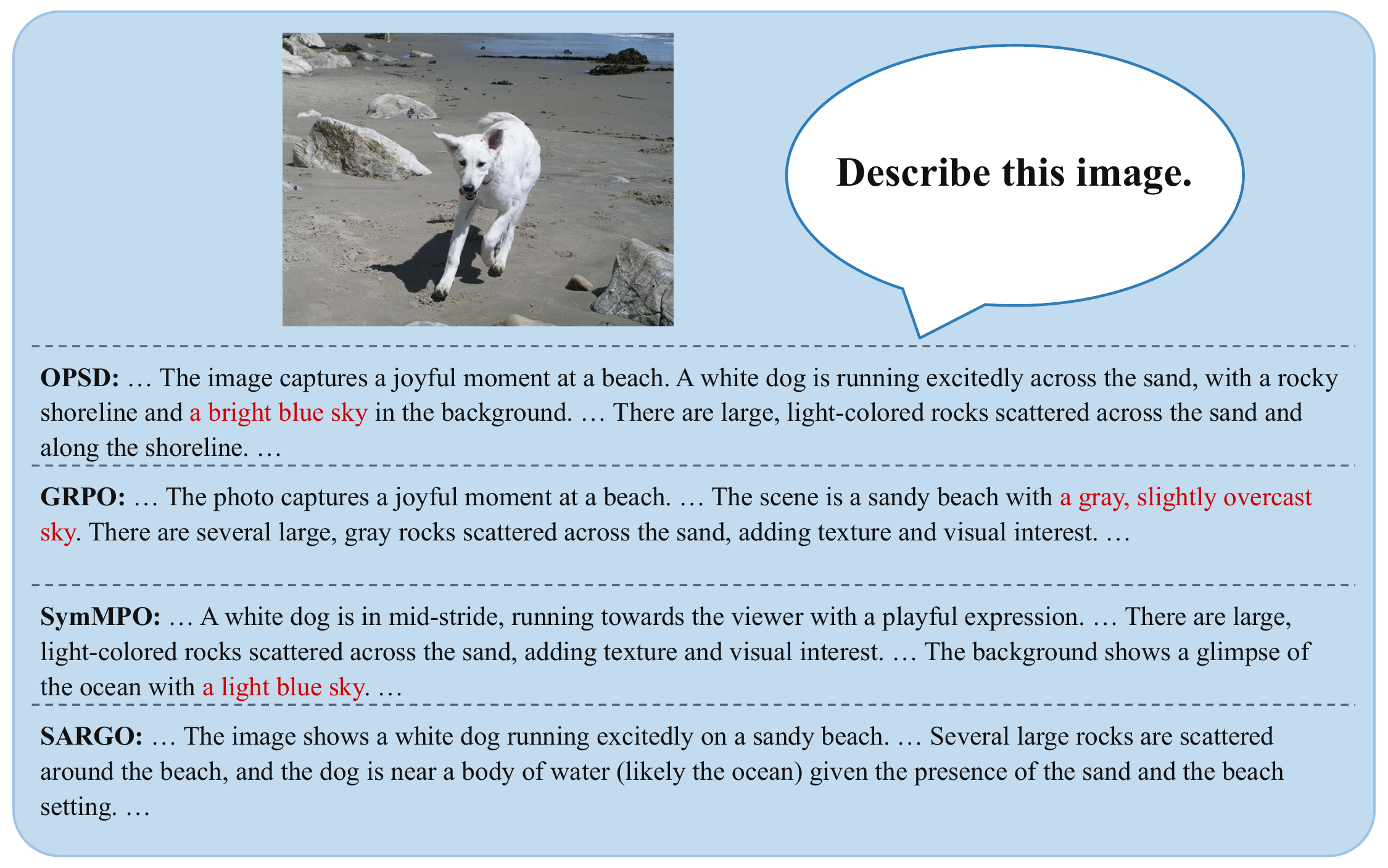}
    \par\medskip
    \includegraphics[
        page=2,
        width=\textwidth,
        height=0.43\textheight,
        keepaspectratio
    ]{figs/case_study.pdf}
    \caption{
        Visualization results comparing OPSD, GRPO, SymMPO, and SARGO.
        Hallucinations are marked in \textcolor{red}{red}.
    }
    \label{fig:case-study-1}
\end{figure*}

We provide qualitative comparisons among OPSD, GRPO, SymMPO, and SARGO on six hallucination examples in Figs.~\ref{fig:case-study-1}--\ref{fig:case-study-3}. These qualitative examples reflect the overall performance gap observed in our quantitative evaluations: compared with SARGO, the baselines produce more hallucinated content, including nonexistent objects and unsupported background details, while SARGO reduces such errors without sacrificing grounded visual information, resulting in descriptions that are both more faithful and informative.

\begin{figure*}[p]
    \centering
    \includegraphics[
        page=3,
        width=\textwidth,
        height=0.43\textheight,
        keepaspectratio
    ]{figs/case_study.pdf}
    \par\medskip
    \includegraphics[
        page=4,
        width=\textwidth,
        height=0.43\textheight,
        keepaspectratio
    ]{figs/case_study.pdf}
    \caption{
        Visualization results comparing OPSD, GRPO, SymMPO, and SARGO.
        Hallucinations are marked in \textcolor{red}{red}.
    }
    \label{fig:case-study-2}
\end{figure*}

\begin{figure*}[p]
    \centering
    \includegraphics[
        page=5,
        width=\textwidth,
        height=0.43\textheight,
        keepaspectratio
    ]{figs/case_study.pdf}
    \par\medskip
    \includegraphics[
        page=6,
        width=\textwidth,
        height=0.43\textheight,
        keepaspectratio
    ]{figs/case_study.pdf}
    \caption{
        Visualization results comparing OPSD, GRPO, SymMPO, and SARGO.
        Hallucinations are marked in \textcolor{red}{red}.
    }
    \label{fig:case-study-3}
\end{figure*}
\section{Impact Statement}
This work aims to improve the reliability of vision-language models by reducing unsupported claims while preserving useful visual information. More faithful multimodal responses can support applications such as visual assistance and image-grounded information access, where hallucinated content may mislead users.

\end{document}